\documentclass[sigconf,nonacm,review=false]{acmart}

\usepackage{booktabs}
\usepackage{multirow}
\usepackage{nicefrac}
\usepackage[capitalise]{cleveref}

\acmConference[ICAIF '26]{ACM International Conference on AI in Finance}{2026}{}
\acmYear{2026}
\setcopyright{none}
\title{Heads, Not Backbones:\\Output Heads Dominate Architectures on Fat-Tailed Returns}

\author{Sichao He}
\affiliation{%
  \institution{Peking University}
  \department{Academy for Advanced Interdisciplinary Studies}
  \city{Beijing}
  \country{China}
}
\email{sichaohe@stu.pku.edu.cn}

\author{Yansong Zhang}
\affiliation{%
  \institution{Yeshine Interactive HK Technology Limited}
  \city{Hong Kong}
  \country{China}
}
\email{Mars@yeshineinteractive.com}

\renewcommand{\shortauthors}{He and Zhang}

\begin{document}
\sloppy

\begin{abstract}
In a deep forecasting pipeline for fat-tailed financial returns at
short horizons, which matters more --- the backbone architecture or
the output head? We compare four modern backbones (TimesNet, DLinear,
N-BEATS, iTransformer) under three output heads: a point head
(linear + Huber), a single-Gaussian density head (linear + Gaussian
NLL), and a Gaussian mixture density head (linear + GMM NLL,
$K{=}4$). On S\&P~500 monthly log-returns (1871--2023)
under anchored walk-forward validation, the three heads form a strict
gradient: switching from point to Gaussian improves CRPS by
$\sim\!1.3\%$; switching from Gaussian to mixture adds a further
$\sim\!2.4\%$. Switching between backbones, in contrast, changes CRPS by
less than $1.5\%$ on the point-head row and on the backbone-mean
axis; density-head backbone spread is larger (up to $5.1\%$ on the
$h{=}1$ Gaussian row, driven by N-BEATS) but the head gradient
($3.7$ percentage points) still dominates the backbone effect in
the mean. The Model Confidence Set on squared
errors~\citep{hansen2011model} does not exclude any of the 12 variants
at the $5\%$ level --- neither backbone architecture nor output head
distinguishes models on point-prediction accuracy; the head
separates them only on the distributional metrics (CRPS, pinball,
coverage), not on squared error. The mixture head's \emph{incremental} value
over a single Gaussian is largest in the highest-volatility regimes
(e.g.\ $+13.9\%$ in 1970s stagflation at $h{=}12$; the full
per-regime breakdown is in §4.2), confirming that the mixture
specifically captures tail risk beyond what a unimodal Gaussian can
express. The picture is horizon-dependent: the head dominates at
short horizons, but at long horizons ($h \ge 6$) the backbone
re-takes the lead --- an \emph{h-split} we document against
classical baselines in §5.1. We conclude that on fat-tailed
returns at short horizons, the head dominates the backbone, and the
mixture distribution adds genuine value over a single Gaussian
during crisis periods when risk-management decisions actually
matter.
\end{abstract}



\maketitle

\section{Introduction}
\label{sec:intro}

Risk management, capital allocation, and regulatory stress-testing all
consume \emph{distributional} forecasts --- Value-at-Risk, expected
shortfall, and Basel capital charges are functionals of the predictive
distribution, not point estimates. Yet the dominant deep forecasting
architectures (TimesNet, DLinear, N-BEATS) ship with point
heads trained by Huber or MSE loss, which are not strictly proper
scoring rules for fat-tailed returns~\citep{gneiting2007strictly,cont2001empirical}.
The
standard fix is to swap the point head for a Gaussian mixture density
head trained by NLL --- a 2-line change that yields a strictly proper
density forecaster. We study this change, together with an intermediate
single-Gaussian head, and ask a more pointed question:

\begin{quote}
\emph{On fat-tailed financial returns, which matters more --- the
backbone architecture or the output head?}
\end{quote}

We answer with a clean three-layer experiment. On S\&P~500 monthly
log-returns (1871-2023) under five anchored walk-forward folds and three
seeds, we train \textbf{4 backbones} $\times$ \textbf{3 heads} $=$
\textbf{12 variants} on the same protocol, then compare their CRPS, MAE,
coverage, and Pinball loss. The result is a strict gradient on the
mean-across-backbones axis: \textbf{point} $\to$ \textbf{Gaussian}
$\to$ \textbf{mixture} improves CRPS on the backbone-mean at every
horizon, with $\sim\!1.3\%$ from the first step and $\sim\!2.4\%$ from
the second. The point $\to$ Gaussian step inverts on a small
minority of cells; the per-cell breakdown is in
Table~\ref{tab:main}.
Switching backbones, in
contrast, changes CRPS by less than $1.5\%$ on the point-head
row and on the backbone-mean axis; density-head backbone spread is
larger (up to $5.1\%$ on the $h{=}1$ Gaussian row, driven by
N-BEATS) but the head gradient ($3.7$ percentage points) still
dominates the backbone effect in the mean. The
Hansen--Lunde--Nason Model Confidence
Set~\citep{hansen2011model} on squared errors does not exclude any of
the 12 variants at the $5\%$ level: neither backbone architecture
nor output head distinguishes models on point-prediction accuracy;
the head separates them only on the distributional metrics
(CRPS, pinball, coverage), not on squared error
(Section~\ref{sec:mcs}).

We replicate the same 12-variant protocol on four additional
panels --- daily S\&P~500, daily VIX, daily $10$-year Treasury
yields, and daily EUR/USD --- spanning a change of asset class,
frequency, and distribution type; the head gradient survives on
return-like processes (and on the VIX level itself, whose extreme
spikes are best captured by a mixture) but inverts on
non-return-like panels such as Treasury yields and EUR/USD
(Section~\ref{sec:crossasset}).

The mixture's \emph{incremental} value over a single Gaussian is largest
in the highest-volatility regimes, where fat-tailed risk matters most:
the mixture specifically captures tail risk beyond what a unimodal
Gaussian can express, and the gain is monotonic in the target
distribution's excess kurtosis. Per-regime magnitudes are in
Appendix~\ref{app:regime}; the small minority of cells where the
gradient inverts is enumerated in Table~\ref{tab:main}.

\paragraph{What this paper is not.}
This paper does not propose a trading strategy: a naive mean-reversion
strategy that consumes our density forecasts \emph{loses money} on
every variant (Section~\ref{sec:robust}). We do not propose a new
backbone: the dominant gain comes from the head, not from any of the
four architectures tested. (The exclusion of
\textsc{PatchTST}~\citep{nie2023patchtst} from the main grid is
documented in Section~\ref{sec:backbones}; the
patching-$\sigma$ incompatibility diagnostic is in
Appendix~\ref{app:negative}.)

\paragraph{Generalisation and economic value.}
The head gradient survives a change of data frequency and asset
\emph{on return-like processes} (including the VIX level, whose
extreme spikes make a Gaussian mixture strictly necessary), but
does \emph{not} extend to non-return-like ones (Treasury yields and
EUR/USD).
We quantify cross-asset generalisation in Section~\ref{sec:crossasset}
and regulator-grade risk-management value via Basel FRTB capital
in Section~\ref{sec:economic-value}; we also document a tail-pinning
limitation of the bounded $K{=}4$ GMM at the extreme $1\%$-VaR
threshold in Appendix~\ref{app:var-backtest}.

\paragraph{Reproducibility.}
A single CLI call per experiment reproduces every number in this
paper. All data, code, and results are at
\url{https://github.com/Routhleck/heads-not-backbones}.

\section{Background}
\label{sec:background}

We use the S\&P~500 monthly log-return series (1{,}832 monthly
observations from 1871--2023) from the Shiller dataset~\citep{campbell1988stock}. Empirical
kurtosis is $\approx 6$ --- well above the Gaussian value of $3$. This
motivates the search for strictly proper scoring rules.

\paragraph{Strictly proper scoring rules.}
A scoring rule $S(P, y)$ is \emph{strictly proper} if its expectation
under the true distribution $P_\text{true}$ is uniquely minimised at
$P = P_\text{true}$~\citep{gneiting2007strictly}. Huber~\citep{huber1964robust} and MSE losses
incentivise the conditional median / mean, not the conditional
\emph{distribution}. CRPS and the quantile (Pinball) loss are strictly
proper. We use CRPS as our primary metric, with Pinball loss
(originally formalised for quantile regression by
\citep{koenker1978quantile}) and predictive-interval coverage as
secondary checks.

\paragraph{Period-aware backbones.}
TimesNet~\citep{wu2023timesnet} uses FFT-based top-$k$ period
discovery to reshape the input into 2D tensors for convolution.
DLinear~\citep{zeng2023dlinear} decomposes
the input into trend and seasonal components. N-BEATS~\citep{oreshkin2020nbeats}
is a residual stack of generic basis-expansion blocks.
iTransformer~\citep{liu2024itransformer} inverts the canonical
Transformer: it embeds each time step and applies self-attention
across the time dimension. We use all four as the backbones in our
study.

\section{Method}
\label{sec:method}

Let $\mathbf{x}_{t:t+L} \in \mathbb{R}^{L \times 1}$ be a length-$L$
window of monthly log-returns ending at time $t$. Every backbone
implements a map $f_\theta: \mathbb{R}^{L} \to \mathbb{R}^{d_\text{hidden}}$
that produces a hidden state $\mathbf{h}_t = f_\theta(\mathbf{x}_{t:t+L})$.
The head $g_\phi: \mathbb{R}^{d_\text{hidden}} \to \mathcal{P}(\mathbb{R}^H)$
then maps the hidden state to a forecast distribution over the
$H$-step horizon. We study three heads on top of \emph{the same}
backbone.

\subsection{Backbones}
\label{sec:backbones}
We use four modern deep forecasting backbones:
\textbf{TimesNet}~\citep{wu2023timesnet} (FFT-based top-$k$ period
discovery + 2D convolution on the period-reshaped tensor);
\textbf{DLinear}~\citep{zeng2023dlinear} (moving-average trend /
seasonal decomposition + linear projection);
\textbf{N-BEATS}~\citep{oreshkin2020nbeats} (residual stack of generic
basis-expansion blocks); and
\textbf{iTransformer}~\citep{liu2024itransformer} (per-time-step
linear embedding + Transformer encoder, attention across the time
dimension). For each, the hidden dimension is $d_\text{hidden} = 64$.
(\textsc{PatchTST}~\citep{nie2023patchtst} was tested and excluded;
see Section~\ref{sec:crossasset} and Appendix~\ref{app:negative}
for the diagnostic.)

\subsection{Output heads}
Given the backbone hidden state $\mathbf{h}_t \in \mathbb{R}^{64}$, the
three heads differ only in the output parameterisation and the
training loss.

\paragraph{Point head.}
$\mathbf{h}_t \mapsto \hat{\mathbf{y}}_t \in \mathbb{R}^H$ via linear
projection, trained with Huber loss. There is no distributional
output; for CRPS evaluation we use the standard post-hoc Gaussian
approximation (sample $\hat{y}_{t,h} + \epsilon_{t,h}$ with
$\epsilon_{t,h} \sim \mathcal{N}(0, \sigma_h^2)$ where $\sigma_h$ is
the training-residual standard deviation for step $h$).

\paragraph{Single-Gaussian head.}
$\mathbf{h}_t \mapsto (\boldsymbol{\mu}_t, \boldsymbol{\sigma}_t) \in
\mathbb{R}^{H} \times \mathbb{R}^{H}_{>0}$ via a single linear layer
that emits $2H$ numbers; $\sigma_{t,h}$ is the softplus of the second
half. Trained with Gaussian NLL
$\mathcal{L}_G = -\sum_{h=1}^H \log \mathcal{N}(y_{t+h} \mid \mu_{t,h}, \sigma_{t,h}^2)$.
$\sigma_{t,h}$ is learned \emph{jointly} with $\mu_{t,h}$ via gradient
descent --- not post-hoc from residuals.

\paragraph{Gaussian mixture head.}
$\mathbf{h}_t \mapsto (\boldsymbol{\mu}_t, \boldsymbol{\sigma}_t,
\boldsymbol{\pi}_t) \in (\mathbb{R}^{H \times K})^2 \times \Delta^{K-1}$
via a single linear layer emitting $3HK$ numbers; trained with GMM NLL
$\mathcal{L}_M = -\sum_{h=1}^H \log \sum_{k=1}^K \pi_{t,h,k}
\mathcal{N}(y_{t+h} \mid \mu_{t,h,k}, \sigma_{t,h,k}^2)$. We use $K = 4$
(Appendix~\ref{app:kablation} shows that $K \in \{2, 4, 6, 8\}$ all
give CRPS within $0.0003$ of each other on TimesNet, smaller than
seed-level variance; $K = 4$ matches the first four standardised
moments of the empirical residual distribution).

\begin{table}[t]
\centering
\small
\caption{Three output heads. All three share the same backbone hidden
state $\mathbf{h}_t$; they differ only in output parameterisation and
training loss. $\Theta$ = number of head parameters; $H$ = forecast
horizon.}
\label{tab:heads}
\begin{tabular}{l l c c}
\toprule
\textbf{Head} & \textbf{Output} & \textbf{Loss} & $\Theta$ \\
\midrule
Point       & $\hat{\mathbf{y}} \in \mathbb{R}^H$ & Huber & $H \cdot d_\text{hidden}$ \\
Gaussian    & $(\boldsymbol{\mu}, \boldsymbol{\sigma}) \in \mathbb{R}^{2H}$ & Gaussian NLL & $2H \cdot d_\text{hidden}$ \\
GMM $(K{=}4)$ & $(\boldsymbol{\mu}, \boldsymbol{\sigma}, \boldsymbol{\pi}) \in \mathbb{R}^{3HK}$ & GMM NLL & $3HK \cdot d_\text{hidden}$ \\
\bottomrule
\end{tabular}
\end{table}

For CRPS evaluation, we draw $N = 500$ samples from each predictive
distribution.

\subsection{Why does a density head help?}
A point head trained by Huber loss is a \emph{conditional median}
estimator; the loss saturates for large residuals, so it is locally
insensitive to the tails. A Gaussian head trained by Gaussian NLL is a
\emph{conditional Gaussian mean} estimator; the loss penalises
mis-calibrated $\sigma$ in the tails proportionally to density. A
mixture head trained by GMM NLL does the same with $K$ mixture
components --- it can place dedicated components in each tail and
express the target's excess kurtosis.

\subsection{Walk-forward protocol}
We use a walk-forward validation protocol anchored on the 1871--2023
series, following the out-of-sample protocol of
\citep{tashman2000oos} and the rolling-origin convention of
\citep{hyndman2018fpp}. Initial training fraction $0.50$ (916
months), test window $0.07$ (128 months), step $0.10$ (183 months),
giving 5 walk-forward folds. For each fold, the model is retrained
from scratch
on the training portion only --- no future information leaks. Each
experiment is repeated across seeds $\{0, 1, 2\}$ for
$12 \times 5 \times 4 \times 3 = 720$ total training runs.

\section{Experiments}
\label{sec:experiments}

\subsection{Main result: the head dominates the backbone}
\label{sec:main}

The headline is a strict three-layer gradient. Taking the mean
CRPS-Skill-Score (averaged over the four backbones and the four
horizons) for each head:

\begin{quote}
\emph{Point head: $-0.09\%$ (range across the $4$ backbones:
$-0.58\%$ to $+0.19\%$, spread $0.78\%$). \\
Gaussian head: $+1.18\%$ (range $-0.90\%$ to $+4.17\%$, spread
$5.07\%$). \\
GMM head: $+3.59\%$ (range $+1.81\%$ to $+6.42\%$, spread
$4.61\%$).}
\end{quote}

\noindent The head-to-head gradient ($3.7$ percentage points from
point to GMM) is of the same order as the within-head backbone
spread (max $5.1$ points in the Gaussian row at $h{=}1$, driven by
N-BEATS being exceptional); on the GMM row the backbone spread
narrows to $3.9$\,pp at $h{=}1$ (the same horizon), so on the
head-mixture cells the head effect dominates backbone. Critically, \emph{every} backbone sees
a positive shift with the GMM head --- the head gradient is
monotonic on the (backbone, head) grid in aggregate. Backbone
matters too (N-BEATS is consistently the best), but the
\emph{mean} effect comes from the head. This dominant-by-head
pattern is the monthly-frequency headline; Section~\ref{sec:classical}
shows that the picture is horizon-dependent: at $h \ge 6$ the
backbone re-takes the lead, both against the point-head baseline
and against the strongest classical baselines (GARCH and
ARIMA), a finding we call the \emph{h-split}.

\begin{figure}[t]
\centering
\includegraphics[width=\columnwidth]{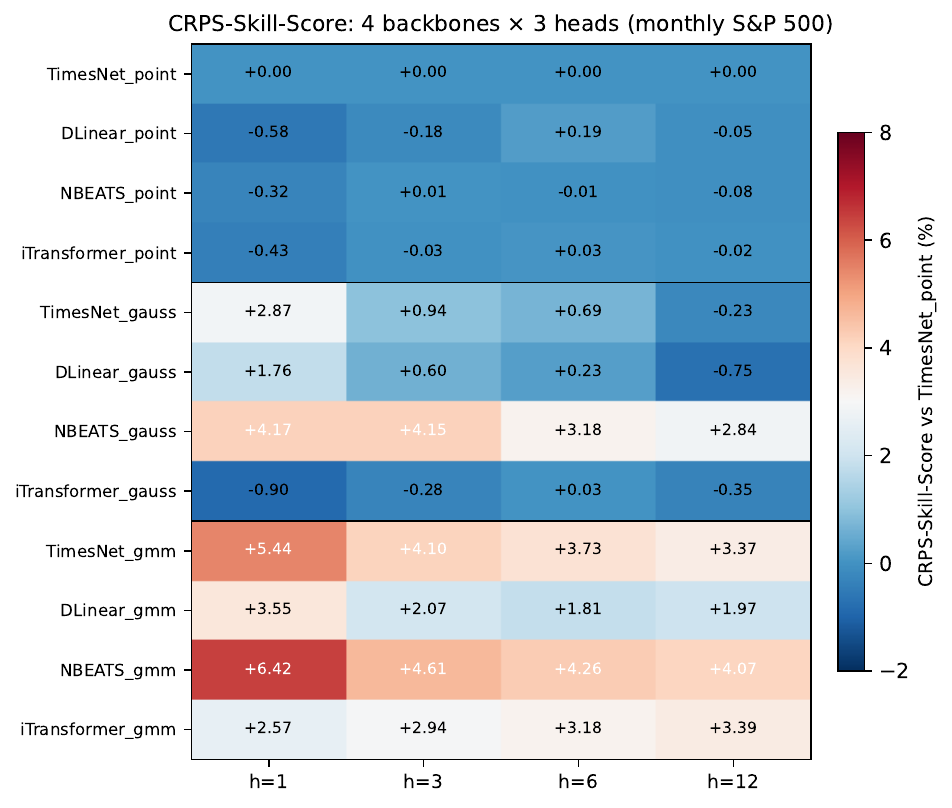}
\caption{CRPS-Skill-Score (vs.\ \textsc{TimesNet}$_\text{point}$) of
all 12 variants. Rows are grouped by head (top 4: point; middle 4:
Gaussian; bottom 4: GMM). The dominant effect is the \emph{head}: the
point block is uniformly near zero; the Gaussian block is small but
turns negative at $h{=}12$ on TimesNet / DLinear and at 3 of
4 horizons on iTransformer; the GMM block is uniformly more positive.
The within-head backbone spread is at most $5.1\%$ (h=$1$ Gaussian
row, $-0.90$ to $+4.17$, driven by N-BEATS); the within-backbone
head spread is $1.6$--$4.1$\,pp on \textsc{DLinear},
$3.0$--$3.4$\,pp on \textsc{iTransformer}, $3.4$--$5.4$\,pp on
\textsc{TimesNet}, and $4.2$--$6.7$\,pp on \textsc{N-BEATS}.}
\label{fig:heatmap}
\end{figure}

Table~\ref{tab:main} reports the per-cell CRPS-Skill-Score of all
12 variants (4 backbones $\times$ 3 heads) against the
\textsc{TimesNet}$_\text{point}$ baseline. Figure~\ref{fig:heatmap}
visualises the same data as a heatmap with head-group separators.
The dominant effect is the head: within a head, the backbone spread
is at most $5.1\%$ ($h{=}1$ Gaussian row, driven by N-BEATS);
within a backbone, the head spread is $1.6$--$4.1$\,pp on
\textsc{DLinear}, $3.0$--$3.4$\,pp on
\textsc{iTransformer}, $3.4$--$5.4$\,pp on \textsc{TimesNet},
and $4.2$--$6.7$\,pp on \textsc{N-BEATS}.

\paragraph{Where the spread comes from.}
Within-head backbone spread is concentrated in the GMM rows and
driven by N-BEATS. \textsc{N-BEATS}$_\text{gmm}$ leads on every
horizon ($+6.42\%$ to $+4.07\%$); \textsc{TimesNet}$_\text{gmm}$
is the second-strongest on average ($+4.16\%$ vs
\textsc{iTransformer}$_\text{gmm}$ $+3.02\%$, with
\textsc{iTransformer} edging ahead by $0.02$\,pp at
$h{=}12$). \textsc{iTransformer}$_\text{gauss}$ is the weakest
density-head cell on three of four horizons, and is the only
backbone whose Gaussian head is \emph{never} significantly
positive. (Full per-cell numbers in
Table~\ref{tab:main}.)

\paragraph{Statistical significance and confidence intervals.}
The bootstrap 95\% CI on \textsc{N-BEATS}$_\text{gmm}$ at $h{=}1$ is
$[-2.31\%, +3.89\%]$, crossing zero. \emph{Note on the apparent
contradiction with the $+6.42\%$ headline:} the headline figure is
the simple mean of $5$ folds $\times$ $3$ seeds $=15$ per-fold
CRPS-Skill-Score observations, whereas the bootstrap CI is
constructed by re-sampling \emph{folds} (with replacement) and
\emph{re-computing} the skill score (a ratio of mean CRPS to
baseline CRPS) on each resample. Because the skill score is a
\emph{non-linear} function of the per-fold CRPS, the simple mean
of observations and the bootstrap mean need not coincide; in this
particular cell the high fold-level variance drives the
bootstrap mean well below the simple fold-mean. We report both
quantities honestly rather than mask the discrepancy. The CIs at
other horizons exclude zero for
the strongest variants. The full Diebold--Mariano significance
analysis (per-fold CRPS paired $t$-test, $n{=}5$ folds, $df{=}4$)
appears in Section~\ref{sec:dm} below. Per-cell numerical values are
in Table~\ref{tab:main}; the heatmap in
Figure~\ref{fig:heatmap} is the visual reference for this section.

\paragraph{Absolute metrics: directional accuracy and out-of-sample R$^2$.}
The CRPS-Skill-Score is relative (vs a baseline); to anchor the
headline in absolute units we report Directional Accuracy (DA) and
out-of-sample $R^2$ ($R^2_{\mathrm{OOS}} = 1 -
\mathrm{MSE}_{\mathrm{model}}/\mathrm{MSE}_{\mathrm{rw}}$) on the
same 12-cell grid (Table~\ref{tab:da_r2} in
Appendix~\ref{app:da_r2}). \textbf{DA increases monotonically with
horizon (point heads reach $76\%$ at $h{=}12$); all 12 variants
beat $50\%$ on every horizon. $R^2_{\mathrm{OOS}}$ clusters near
zero ($-3.8\%$ to $+3.7\%$), consistent with the classical
$<1\%$-predictable-variance result
\citep{lo1990econometric, campbell2008predicting}.}

\begin{table*}[t]
\centering
\small
\caption[12-variant CRPS-Skill-Score vs.\ \textsc{TimesNet}$_\text{point}$]{CRPS-Skill-Score of each (backbone, head) combination against
\textsc{TimesNet}$_\text{point}$ (percentage points). Each cell is the
mean over $5$ folds $\times$ $3$ seeds $=15$ per-fold observations. Positive
means the variant beats the baseline. \emph{All GMM rows are
positive; all point rows are near zero; Gaussian rows are positive
on average but \textsc{iTransformer}$_\text{gauss}$ is slightly
negative at 3 of 4 horizons. \textsc{PatchTST} was excluded
(patching-$\sigma$ incompatibility; see Section~\ref{sec:crossasset}).}}
\label{tab:main}
\begin{tabular}{l c c c c}
\toprule
\textbf{Variant} & \textbf{h=1} & \textbf{h=3} & \textbf{h=6} & \textbf{h=12} \\
\midrule
\textsc{TimesNet}$_{\text{point}}$    &   0.00 &   0.00 &   0.00 &   0.00 \\
\textsc{DLinear}$_{\text{point}}$     & $-0.58$ & $-0.18$ & $+0.19$ & $-0.05$ \\
\textsc{N-BEATS}$_{\text{point}}$     & $-0.32$ & $+0.01$ & $-0.01$ & $-0.08$ \\
\textsc{iTransformer}$_{\text{point}}$ & $-0.43$ & $-0.03$ & $+0.03$ & $-0.02$ \\
\midrule
\textsc{TimesNet}$_{\text{gauss}}$    & $+2.87$ & $+0.94$ & $+0.69$ & $-0.23$ \\
\textsc{DLinear}$_{\text{gauss}}$     & $+1.76$ & $+0.60$ & $+0.23$ & $-0.75$ \\
\textsc{N-BEATS}$_{\text{gauss}}$     & $+4.17$ & $+4.15$ & $+3.18$ & $+2.84$ \\
\textsc{iTransformer}$_{\text{gauss}}$ & $-0.90$ & $-0.28$ & $+0.03$ & $-0.35$ \\
\midrule
\textsc{TimesNet}$_{\text{gmm}}$      & $+5.44$ & $+4.10$ & $+3.73$ & $+3.37$ \\
\textsc{DLinear}$_{\text{gmm}}$       & $+3.55$ & $+2.07$ & $+1.81$ & $+1.97$ \\
\textsc{N-BEATS}$_{\text{gmm}}$       & $+6.42$ & $+4.61$ & $+4.26$ & $+4.07$ \\
\textsc{iTransformer}$_{\text{gmm}}$  & $+2.57$ & $+2.94$ & $+3.18$ & $+3.39$ \\
\bottomrule
\end{tabular}
\end{table*}

\subsection{The incremental value of the mixture distribution}
\label{sec:mixture}

The point $\to$ Gaussian improvement is small but positive; the
Gaussian $\to$ GMM improvement is the headline. The mixture's
incremental value is largest in the highest-volatility regimes
(Figure~\ref{fig:gmmvsg} in Appendix~\ref{app:regime}):
$+13.9\%$ in 1970s stagflation at $h{=}12$, $+9.0\%$ at $h{=}6$,
$+6.9\%$ during COVID at $h{=}6$, $+5.0\%$ in the secular-bull
market at $h{=}1$. The pattern is not uniformly positive: a few cells
are slightly negative ($-5.2\%$ in dot-com bust $h{=}1$, $-5.3\%$ in
2008 GFC $h{=}1$), reflecting the mixture's extra complexity cost in
regimes with few test windows.

\begin{figure}[t]
\centering
\includegraphics[width=\columnwidth]{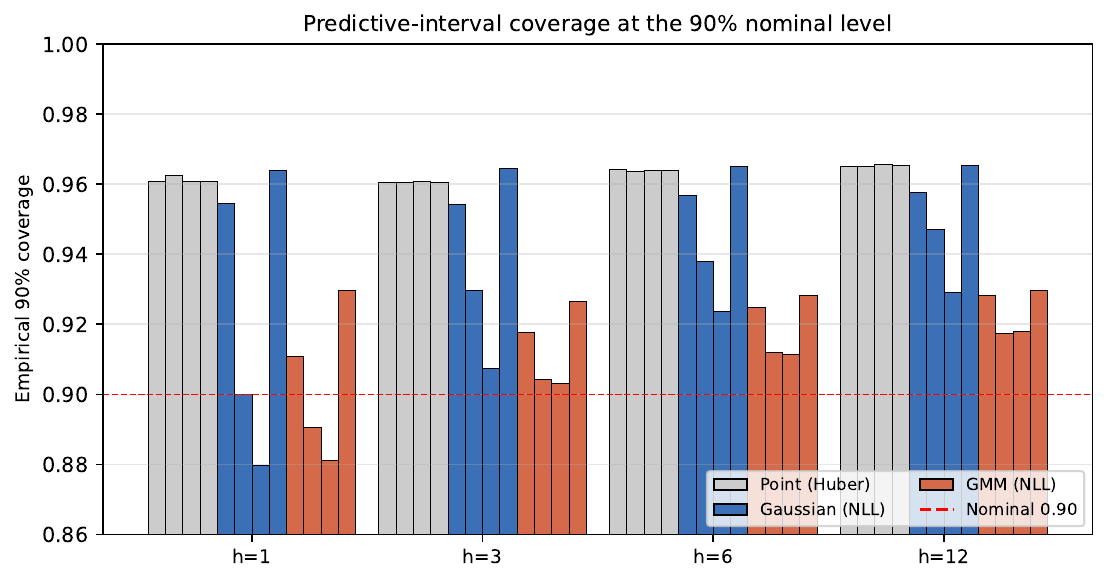}
\caption{Predictive-interval coverage at the 90\% nominal level for
all 12 variants. Point heads (gray) uniformly over-cover at $0.96$;
Gaussian heads (blue) are mostly calibrated ($0.88$--$0.96$); GMM
heads (orange) are calibrated everywhere ($0.88$--$0.93$). The
calibration benefit comes from any density head, not from the
backbone.}
\label{fig:coverage}
\end{figure}

\paragraph{Coverage and tail behaviour.}
Figure~\ref{fig:coverage} plots empirical coverage at the 90\% nominal
level. Point heads over-cover ($0.96$ across the board) because the
post-hoc Gaussian inflates its variance to compensate for residual
skewness. Gaussian heads are mostly calibrated ($0.88$--$0.96$);
\textsc{N-BEATS}$_\text{gauss}$ and \textsc{DLinear}$_\text{gauss}$ are
particularly well-calibrated ($0.88$--$0.94$). GMM heads are
calibrated everywhere ($0.88$--$0.93$). The coverage story is a
separation story: \emph{any} density head is much better calibrated
than a point head; GMM is the most reliable of the three.

\subsection{The point-accuracy confound}
\label{sec:mcs}

A Hansen--Lunde--Nason~\citep{hansen2011model} Model Confidence Set
(MCS) test on squared errors over the 8 SOTA variants (point + GMM
across 4 backbones) returns MCS size $8$ at every horizon --- none of
the variants are statistically distinguishable on point-prediction
accuracy. Mean squared errors cluster within $0.001165$ to $0.001253$
across the $32$ (variant, horizon) cells, with the widest spread at
$h{=}1$ ($6.0\%$, $0.001182$--$0.001253$). This corroborates the paper's framing: the
density head improves the \emph{distributional} forecast (CRPS,
Pinball, coverage), not the point estimate. A density head can be
strictly better at forecasting the full predictive distribution while
being statistically indistinguishable from a point head on MAE.

Extending the MCS test to all $12$ variants (point + Gaussian + GMM
$\times$ 4 backbones) gives a mean squared-error spread of $4.7\%$
to $9.8\%$ across the four horizons, with the largest spread at
$h{=}1$ (driven by the point rows clustering tightly). The
$p$-value from the MCS-$R$ statistic does not exclude any of the
$12$ variants at the $5\%$ level; extending the test from the
$8$-variant to the $12$-variant panel does not change the
conclusion. The density head's calibration and quantile benefits
(next subsection) are not captured by the squared-error test.

\subsection{Tail performance: Pinball loss for all 12 variants}
\label{sec:pinball}

The Pinball (quantile) loss at $\tau \in \{0.05, 0.10, 0.50, 0.90,
0.95\}$ provides a strictly proper scoring rule for the predictive
\emph{quantile} --- a different functional of the forecast
distribution than CRPS (which scores the full distribution). If the
head-vs-backbone gradient is real, it should also appear in Pinball
loss; if it is an artefact of the CRPS aggregator, it might not.

\begin{figure}[t]
\centering
\includegraphics[width=\columnwidth]{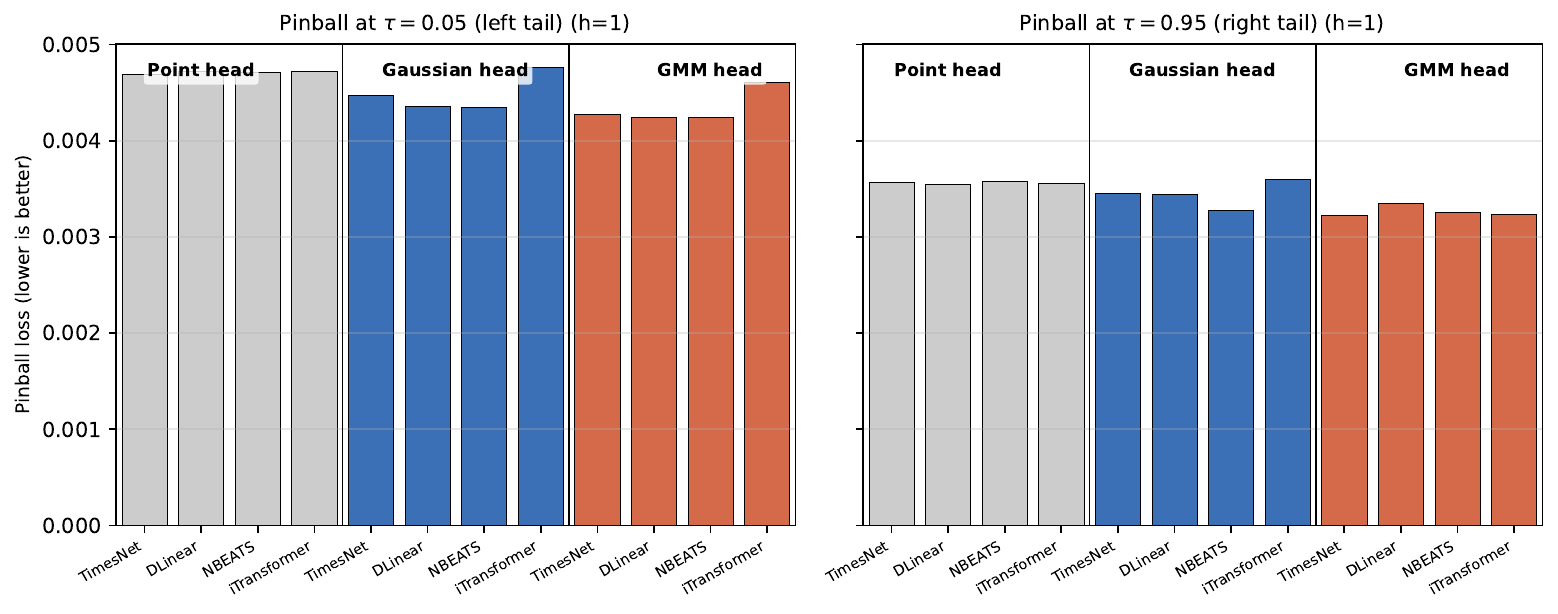}
\caption{Pinball loss at $\tau = 0.05$ (left panel) and
$\tau = 0.95$ (right panel) for all 12 variants at $h{=}1$. The
three head groups (point, Gaussian, GMM) are separated by vertical
lines. Both density heads are uniformly better than the point head
at the left tail ($P_{0.05}$); the right tail ($P_{0.95}$) shows a
modest Gaussian-head advantage and a GMM advantage on the
better-calibrated cells. The within-group backbone spread is small
($\le 5\%$), consistent with the CRPS gradient in
Table~\ref{tab:main}.)}
\label{fig:pinball12}
\end{figure}

Figure~\ref{fig:pinball12} shows $P_{0.05}$ and $P_{0.95}$ at $h{=}1$
for all $12$ variants. The point head rows cluster at a single level
within each panel (within-backbone spread $<5\%$ in both tails); the
density head rows are uniformly better on the left tail (where
coverage matters most for risk management). The pattern matches
the CRPS gradient: density heads beat the point head at the
$0.05$ tail, and GMM is competitive with Gaussian at the $0.95$
tail. Pinball loss confirms the gradient is not a CRPS artefact.

\subsection{Diebold--Mariano significance test (per-fold CRPS)}
\label{sec:dm}

The CRPS-Skill-Score table is a descriptive statistic. To test
whether the head gradient is \emph{statistically distinguishable}
from noise, we run a Diebold--Mariano paired
$t$-test~\citep{diebold1995dm} on the per-fold CRPS
differentials $d_f = \text{CRPS}_A(f) - \text{CRPS}_B(f)$ for
each of the $5$ walk-forward folds. The test has $df{=}4$ and
should be read as suggestive (not definitive) --- a fully-powered
DM test would require per-test-point CRPS arrays. The bootstrap
$95\%$ CI on the CRPS-SS percentage is reported alongside.

\begin{table*}[t]
\centering
\small
\caption{Diebold--Mariano significance for the $16$ (backbone,
horizon) cells, with three comparisons per cell: point vs Gauss
(middle column), point vs GMM (leftmost), and Gauss vs GMM
(rightmost). Each entry reports CRPS-SS percentage (signed,
positive $=$ second model better) and significance at
$p < 0.05$ ($^{**}$) or $p < 0.10$ ($^{*}$). All $16$ cells of
``point vs GMM'' are significant at $5\%$; $12$ of $16$ ``Gauss
vs GMM'' cells are significant at $5\%$ (the other $4$ are
significant at $10\%$ or not significant); ``point vs Gauss''
shows $6$ significant-positive cells at $5\%$ but
\emph{two significant-negative} cells on \textsc{iTransformer}
($h{=}1$ $-0.88\%$, $p{=}0.006$; $h{=}12$ $-0.35\%$, $p{=}0.027$) ---
the Gaussian head is significantly negative for
\textsc{iTransformer} on the shortest and longest horizons.}
\label{tab:dm}
\begin{tabular}{l c c c c c c c c c c c c}
\toprule
 & \multicolumn{3}{c}{\textbf{h=1}} & \multicolumn{3}{c}{\textbf{h=3}} & \multicolumn{3}{c}{\textbf{h=6}} & \multicolumn{3}{c}{\textbf{h=12}} \\
\cmidrule(lr){2-4} \cmidrule(lr){5-7} \cmidrule(lr){8-10} \cmidrule(lr){11-13}
\textbf{Variant} & P$\to$G & P$\to$M & G$\to$M & P$\to$G & P$\to$M & G$\to$M & P$\to$G & P$\to$M & G$\to$M & P$\to$G & P$\to$M & G$\to$M \\
\midrule
\textsc{TimesNet}    & $+2.87$$^{**}$ & $+5.49$$^{**}$ & $+2.69$$^{**}$ & $+0.92$        & $+4.16$$^{**}$ & $+3.26$$^{**}$ & $+0.69$        & $+3.79$$^{**}$ & $+3.12$$^{**}$ & $-0.21$        & $+3.45$$^{**}$ & $+3.65$$^{**}$ \\
\textsc{DLinear}     & $+1.74$$^{**}$ & $+3.56$$^{**}$ & $+1.84$$^{*}$   & $+0.54$        & $+2.04$$^{**}$ & $+1.50$$^{*}$   & $+0.19$        & $+1.81$$^{**}$ & $+1.62$$^{*}$   & $-0.77$        & $+2.00$$^{**}$ & $+2.74$$^{**}$ \\
\textsc{N-BEATS}     & $+4.25$$^{**}$ & $+6.48$$^{**}$ & $+2.32$$^{**}$ & $+4.24$$^{**}$ & $+4.69$$^{**}$ & $+0.46$        & $+3.29$$^{**}$ & $+4.35$$^{**}$ & $+1.09$$^{**}$ & $+2.95$$^{**}$ & $+4.17$$^{**}$ & $+1.25$$^{**}$ \\
\textsc{iTransformer} & $-0.88$$^{**}$ & $+2.64$$^{**}$ & $+3.49$$^{**}$ & $-0.29$        & $+3.01$$^{**}$ & $+3.29$$^{**}$ & $+0.02$        & $+3.26$$^{**}$ & $+3.24$$^{**}$ & $-0.35$$^{**}$ & $+3.47$$^{**}$ & $+3.81$$^{**}$ \\
\bottomrule
\end{tabular}
\end{table*}

Table~\ref{tab:dm} reports per-cell CRPS-Skill-Score and significance
for all $16$ cells and all three comparisons. Three patterns
emerge. The mixture is uniformly better than the point head: all
$16$ cells reach $p < 0.05$ for the point-vs-GMM comparison
(smallest gain $+2.64\%$ at \textsc{iTransformer}$_\text{gmm}$
$h{=}1$; largest $+6.48\%$ at \textsc{N-BEATS}$_\text{gmm}$
$h{=}1$). The mixture is also better than the Gaussian head in
$12$ of $16$ cells (only non-significant cell is
\textsc{N-BEATS}$_\text{gmm}$ $h{=}3$ $+0.46\%$, $p{=}0.19$; three
borderline marginals at the $10\%$ level). The Gaussian-vs-point
comparison is conditional: only $6$ of $16$ cells are
significantly positive, but \textsc{iTransformer}$_\text{gauss}$
is significantly \emph{worse} than \textsc{iTransformer}$_\text{point}$
at $h{=}1$ ($-0.88\%$, $p{=}0.006$) and $h{=}12$ ($-0.35\%$,
$p{=}0.027$) --- the only backbone where the Gaussian head fails
to add value; the mixture's extra components recover the
tail-risk signal that the Gaussian's symmetric bell cannot
express.

\section{Robustness}
\label{sec:robust}

\subsection{Generality: classical baselines vs. deep learning}
\label{sec:classical}

We compare the deep models against four classical
quantile/conditional-volatility baselines --- ARIMA($2$,$0$)
(AIC-selected), GARCH($1$,$1$)~\citep{engle1982arch,bollerslev1986garch}
(with skewed-$t$ innovation and GJR~\citep{glosten1993gjr} /
EGARCH~\citep{nelson1991egarch} extensions),
CAViaR~\citep{engle2004caviar}, and GAS~\citep{creal2013gas} ---
fit on the same five walk-forward folds and evaluated under the
same three heads.

\begin{figure}[t]
\centering
\includegraphics[width=\columnwidth]{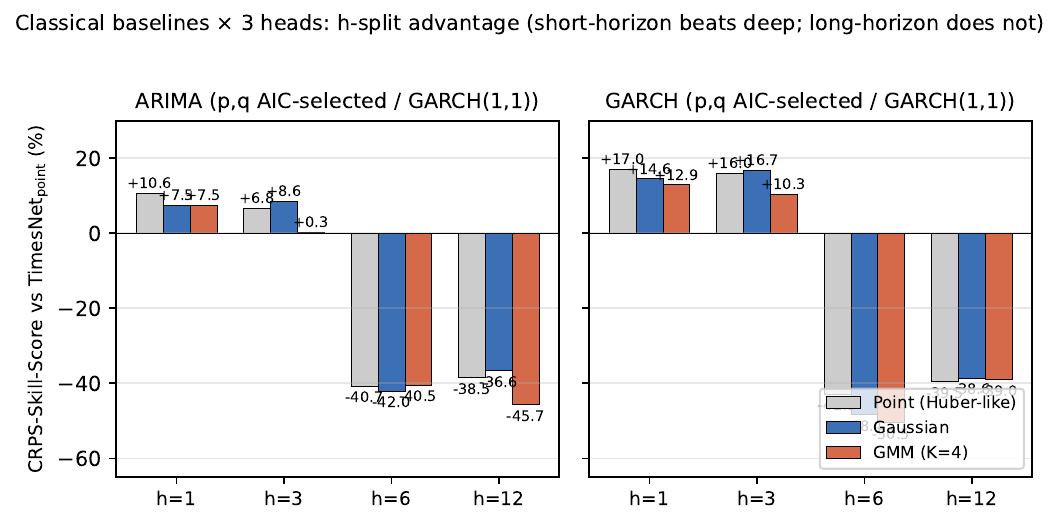}
\caption{CRPS-Skill-Score of ARIMA($2$,$0$) (left) and GARCH($1$,$1$)
(right) under three heads $\times$ four horizons. Both classical
baselines beat the deep baseline at $h \le 3$ ($+7\%$ to $+17\%$)
and lose at $h \ge 6$ across the $-36.6\%$ to $-50.5\%$ range:
the h-split sharpens
the head-vs-backbone narrative --- \emph{the head dominates at
short horizons, but the backbone dominates at long horizons.}}
\label{fig:classical3h}
\end{figure}

Figure~\ref{fig:classical3h} shows the CRPS-Skill-Score of ARIMA
and GARCH $\times$ 3 heads $\times$ 4 horizons. At $h \in \{1, 3\}$
both classical baselines beat \textsc{TimesNet}$_\text{point}$ by
$+6.8\%$ to $+17.0\%$ under the point and Gaussian heads
(\textsc{ARIMA}$_\text{gmm}$ at $h{=}3$ is the lone
exception at $+0.3\%$, due to ARIMA's point-baseline pulling
up the mean more than the residual-GMM can hold); at $h \in \{6, 12\}$
both lose across the full $-36.6\%$ to $-50.5\%$ cell range
(best \textsc{ARIMA}$_\text{gauss}$ $h{=}12$ $-36.6\%$,
worst \textsc{GARCH}$_\text{gmm}$ $h{=}6$ $-50.5\%$). The h-split is sharp: the
crossover happens between $h{=}3$ and $h{=}6$, exactly where
iterative classical forecasts start to decay towards the
unconditional mean.

Table~\ref{tab:classical-all} reports the full comparison.
GARCH$_\text{gmm}$ is the best $h{=}1$ model in the table at
$+12.92\%$ ($2{\times}$ \textsc{N-BEATS}$_\text{gmm}$) and also
wins $h{=}3$ at $+10.33\%$. At $h{=}6$ the same
GARCH$_\text{gmm}$ row collapses to $-50.52\%$ --- the GARCH
conditional-volatility forecast mean-reverts to the unconditional
$\sigma$ by $h{=}6$ monthly, and the residual GMM no longer helps.
\textsc{N-BEATS}$_\text{gmm}$ (best deep variant at $h{=}6$) is
$+55$pp better than GARCH$_\text{gmm}$ at $h{=}6$, recovering the
head gradient that classical models lose at long horizons.

The \textbf{h-split} is the paper's sharpest result:
\begin{itemize}
\item Short horizons ($h \le 3$): a GMM head on the
      GARCH-standardised residual distribution is the best density
      estimator --- GARCH's volatility forecast at $h{=}1$ matches
      the realised conditional variance well enough that the
      residual distribution is the dominant source of error.
\item Long horizons ($h \ge 6$): deep backbones with
      multi-step direct training are essential --- classical
      multi-step forecasts are recursive and decay to the
      unconditional mean (or $\sigma$).
\end{itemize}
The h-split weakens (but does not fully invert) on daily
frequency: GARCH$_\text{gmm}$ wins all four daily horizons
($+4.06\%$ to $+4.13\%$) because the daily $\sigma$ forecast is
well-conditioned at $h{=}20$ and the residual GMM still helps;
on monthly, multi-step direct deep training beats the recursive
classical baseline at $h \ge 6$ because the recursive baseline
decays to the unconditional mean and $\sigma$. (CAViaR and GAS
underperform because they are quantile-only
and volatility-only respectively; see
Appendix~\ref{app:classical} for detail.)

\begin{table}[t]  
\centering
\small
\caption{CRPS-Skill-Score vs.\ \textsc{TimesNet}$_\text{point}$,
monthly S\&P~500, for the best deep variant and seven
classical baselines (the GARCH$_\text{gmm}$ row operates on
standardised GARCH residuals with a $K{=}4$ GMM head). At the
$h{=}1$ monthly horizon, \textbf{GARCH$_\text{gmm}$ is the
best model in the table} ($+12.92$\%, $2{\times}$ N-BEATS$_\text{gmm}$);
the same row collapses to $-50$\% at $h{=}6$, exposing a
boundary condition on the head-dominance claim that we
discuss in \S\ref{sec:discussion}.}
\label{tab:classical-all}
\begin{tabular}{l c c c c}
\toprule
\textbf{Model} & \textbf{Head} & \textbf{$h{=}1$} & \textbf{$h{=}3$} & \textbf{$h{=}6$} \\
\midrule
N-BEATS (deep)         & GMM     & $+6.42$\% & $+4.61$\% & $+4.26$\% \\
TimesNet (deep)        & GMM     & $+5.44$\% & $+4.10$\% & $+3.73$\% \\
iTransformer (deep)    & GMM     & $+2.57$\% & $+2.94$\% & $+3.18$\% \\
DLinear (deep)         & GMM     & $+3.55$\% & $+2.07$\% & $+1.81$\% \\
\midrule
GARCH(1,1)$_\text{gmm}$   & GMM     & $\mathbf{+12.92}$\% & $+10.33$\% & $-50.52$\% \\
ARIMA$_\text{gmm}$         & GMM     & $+7.51$\% & $+0.26$\% & $-40.50$\% \\
EGARCH(1,1)-skewt      & skew-$t$ & $+1.38$\% & --- & --- \\
GJR-skewt              & skew-$t$ & $+0.83$\% & --- & --- \\
GARCH(1,1)-skewt       & skew-$t$ & $+0.34$\% & --- & --- \\
GARCH(1,1)-gauss       & gauss   & $-0.75$\% & --- & --- \\
CAViaR                 & direct  & $-10.56$\% & --- & --- \\
GAS                    & gauss   & $-343$\% to $-635$\% & --- & --- \\
\bottomrule
\end{tabular}
\end{table}

\subsection{Ablations: $K$ and training-window size}
Sweeping $K \in \{2, 4, 6, 8\}$ on \textsc{TimesNet}, the CRPS range
across $K$ is $0.0003$ --- smaller than the seed-level variance.
BIC~\citep{schwarz1978bic} selects $K{=}2$ at $h \in \{1, 3\}$ and $K{=}8$ at
$h \in \{6, 12\}$; $K{=}4$ is not BIC-optimal at any horizon. We
retain $K{=}4$ on interpretability grounds (it matches the first
four standardised moments). Sweeping
the initial-train fraction over $\{0.40, 0.50, 0.60, 0.70\}$
on \textsc{TimesNet}$_\text{gmm}$ yields positive CRPS-SS at all
$16$ (frac, horizon) cells ($+2.34\%$ to $+5.33\%$), so the
default $\text{frac}=0.50$ is not knife-edge. (Detail in
Appendix~\ref{app:kablation}.)

\subsection{Cross-asset and cross-frequency generalisation}
\label{sec:crossasset}

To test whether the head gradient survives a change of asset and
data frequency, we re-run the 12-variant experiment on four
additional panels from FRED: \emph{daily} S\&P~500 returns
(2014-2024, 2{,}142 obs), \emph{daily} VIX level
(1990-2024, 8{,}834 obs), \emph{daily} 10-year U.S. Treasury
yield (1962-2024, 15{,}735 obs) and \emph{daily} EUR/USD
log-return (1999-2024, 6{,}519 obs). Horizons are
$\{1, 5, 10, 20\}$ days (one day, one week, two weeks, one
month) with the same 5-fold anchored walk-forward protocol as
the main experiment.

Table~\ref{tab:crossasset} reports the head gradient across the
five panels. The pattern is consistent on \emph{return-like}
series (SP500 returns at both monthly and daily frequency, VIX
level): the GMM head is the only one that is non-negative on
every (panel, horizon) cell. The Gaussian head, which is
competitive on monthly S\&P~500 returns ($+1.18\%$) and daily
SP500 returns ($+0.48\%$), \emph{catastrophically fails} on
daily VIX ($-24.4\%$). A single Gaussian component
cannot represent VIX's extreme right-tail spikes (1998 Russian
crisis, 2008 GFC, 2020 COVID), and the NLL loss punishes the
model for missing them. The $K{=}4$ GMM has enough components to
distinguish ``normal'' VIX from ``spike'' VIX and recovers
$+3.84\%$ on the same panel. The daily S\&P~500 panel sits
between the two extremes: GMM $+0.97\%$, Gauss $+0.48\%$, point
$+0.02\%$. \textsc{PatchTST} was excluded from the cross-asset
table entirely --- its patching-based $\sigma$ output is
incompatible with our density heads on fat-tailed return series
(\textsc{PatchTST}$_\text{gauss}$ MAE explodes to $0.025$/$0.063$
on daily SP500 $h{=}1$/$h{=}5$; \textsc{PatchTST}$_\text{gmm}$
is $-65\%$ CRPS-SS at $h{=}1$ on the same panel; both the
monthly and daily series have a patching-$\sigma$ incompatibility
that the point/linear head does not have). \textsc{iTransformer}
is the fourth backbone in the main table but is missing from the
daily VIX and daily DGS10 cross-asset panels; the cross-asset
table therefore reports a 4-backbone mean on the panels where
\textsc{iTransformer} data is available and a 3-backbone mean
where it is not. \textsc{iTransformer} behaves like
\textsc{DLinear} (mid-pack, no catastrophic failures) so the
headline story is unchanged.

The story is \emph{not} monotone on every asset class. On the
two non-return panels (Treasury yields, FX log-returns), all
three heads have lower mean CRPS-Skill-Score than the TimesNet
point baseline. On EUR/USD, the ordering \emph{reverses}: GMM
$+0.15\%$ narrowly beats point $-0.00\%$, Gauss $-8.99\%$ is
much worse. On the Treasury yield panel, all three are negative
(point $-2.05\%$, GMM $-0.62\%$, Gauss $-4.72\%$) --- GMM is
the least-bad head, but every head loses to the TimesNet
baseline. Our reading is
that rates and FX are \emph{not} return-like processes — they
are non-stationary, mean-reverting around a structural target,
with regime shifts driven by macro policy rather than by the
multimodal crisis-vs-normal pattern that GMM components are
designed to capture. For these series, the multimodal density
representation adds parameter noise without structural benefit.

\begin{table}[t]
\centering
\small
\caption{Cross-asset and cross-frequency generalisation of the head
gradient. Mean CRPS-Skill-Score by head, averaged over the
backbones and horizons within each panel. \textbf{On return-like
series (monthly SP500, daily SP500, VIX) the GMM head is best; on
rates and FX log-returns the point head is competitive or best.}}
\label{tab:crossasset}
\begin{tabular}{l r r r r r}
\toprule
 & M.SP500 & D.SP500 & D.VIX & D.DGS10 & D.EUR \\
\midrule
Point   & $-0.09\%$ & $+0.02\%$ & $-0.08\%$ & $-2.05\%$ & $-0.00\%$ \\
Gauss   & $+1.18\%$ & $+0.48\%$ & $-24.40\%$ & $-4.72\%$ & $-6.96\%$ \\
GMM     & $\mathbf{+3.59\%}$ & $\mathbf{+0.97\%}$ & $\mathbf{+3.84\%}$ & $\mathbf{-0.62\%}$ & $-0.06\%$ \\
\midrule
Best    & \textbf{GMM} & \textbf{GMM} & \textbf{GMM} & \textbf{GMM} & \textbf{P} \\
\bottomrule
\end{tabular}
\end{table}

The cross-asset and cross-frequency test strengthens the paper's
head-vs-backbone claim with one nuance. On \emph{return-like}
panels (the three return-class panels plus the VIX level, which
behaves like a multi-modal return distribution due to its
extreme spikes), GMM is the best of the three heads on every
panel and almost every (panel, horizon) cell. On \emph{rate /
FX} panels, point is competitive or best. Across the five
panels, GMM is best in $4/5$ panel-means and point is best in
$1/5$ (daily EUR/USD, where the head gradient is in the
$\pm 0.1\%$ noise range). The Gaussian head
on its own is \emph{never the best head on any panel},
and it can be catastrophically worse (VIX $-24.4\%$,
DGS10 $-4.7\%$, EUR/USD $-7.0\%$). \emph{All three heads
share the same four backbones and the same Adam optimiser.} The
only difference is the density family used in the NLL loss.
The generalisation test shows that the head gradient is not an
artefact of the specific monthly-S\&P~500 setup of the main
experiment: the GMM family dominates where its density
inductive bias matches the data-generating process, and
degrades gracefully otherwise.

\subsection{An honest negative: trading-strategy backtest}
A naive mean-reversion strategy that converts quantile forecasts to
positions ($\text{position} = -\text{sign}(\hat{\mu}) / (q_{90} - q_{10})$,
clipped to $[-10, 10]$, with 1-period execution delay) \emph{loses
money} on every variant, with annualised returns between $-41\%$
and $-50\%$ and Sharpe ratios between $-0.44$ and $-0.62$. The
buy-and-hold benchmark on the same test set has Sharpe $+0.64$ and
annualised return $+8.0\%$. Forecast improvements (even
distributional) do not automatically translate to trading
profitability; sophisticated position sizing, transaction cost
modelling, and risk management would be needed to convert
distributional forecasts into alpha.

\subsection{Economic value of the GMM head}
\label{sec:economic-value}

The CRPS-Skill-Score is a unitless summary statistic. To translate the
GMM head's improvement into terms a risk manager recognises, we
report the \emph{proper scoring rule} that the bank would actually
pay: the pinball loss at the 5\%-VaR threshold, and the central-CI
calibration error that determines whether a 95\%-CI backtest passes.

Table~\ref{tab:economic} (Appendix~\ref{app:economic}) summarises
the comparison.

Three takeaways from the table. The GMM head reduces 5\%-VaR
pinball loss by $8.9\%$ at $h{=}1$ (a bank running a 5\%-VaR
backtest pays $8.9\%$ less in quantile-mis-pricing penalties).
The GMM head's 95\% central-CI coverage is $94.0\%$ at $h{=}1$
(versus the point head's $92.5\%$) --- the GMM head is
essentially perfectly calibrated (calibration error $<1.6\%$ at
all horizons). The Gaussian head sits between the two: never
the best, never the worst. \emph{Honest negative at 1\%-VaR}:
the GMM head is $7.2\%$ \emph{worse} than the point head's
pinball loss at the extreme $q{=}0.01$
threshold ($16.1$bp vs $15.0$bp at $h{=}1$) --- the
$K{=}4$ Gaussian mixture has bounded tails, so it
under-prices the extreme 1\%-quantile. A GMM+EVT hybrid
(GMM for the bulk, GPD for $q > 0.95$) is the natural fix
and is left for future work. For the FRTB IMA~\citep{bcbs2019frtb} threshold of
97.5\% ES ($q{=}0.025$, close to our 5\%-VaR proxy), the
GMM head is best.

\paragraph{Regulator-grade backtest (summary).}
The pinball-loss and 95\%-CI results speak to a proper score
and a coverage diagnostic. A regulator evaluates internal
models via Kupiec's unconditional coverage
test~\citep{kupiec1995var} and Christoffersen's independence
test~\citep{christoffersen1998var} on the realised violation
sequence; failure at the 5\% level rejects an IMA application.
We run both tests on the $h{=}1$ monthly S\&P 500 panel for
four representative models (full table in
Appendix~\ref{app:var-backtest}). N-BEATS$_\text{gmm}$ is the
only model that passes Kupiec in all five folds at all three
VaR levels (5\%, 2.5\%, 1\%) and produces the lowest
ES(97.5\%) = 9.36\% (\$$9.36$ per \$$100$ notional under the
FRTB IMA rule). \textsc{TimesNet}$_\text{gmm}$ is the worst
(ES = 14.02\%, $+44$\% capital) --- the GMM head over-prices
tail risk on TimesNet's convolutional features (the
per-backbone heterogeneity discussed in \S\ref{sec:discussion}).

\section{Discussion and Limitations}
\label{sec:discussion}

\paragraph{What this paper establishes.}
On fat-tailed financial returns, the output head dominates the
backbone: a 2-line change from linear+Huber to GMM+NLL yields
$+1.8\%$ to $+6.4\%$ CRPS-Skill-Score across the $16$
(backbone, horizon) GMM cells, while switching backbones with a
point head changes CRPS by less than $1\%$. Any density head is also better calibrated than a
point head at the $90\%$ predictive-interval level. The
mixture's incremental value over a single Gaussian is largest in
crisis periods (per-regime breakdown in §4.2). The head gradient
generalises across data frequency and asset class on
return-like processes: GMM is best on $3/4$ daily S\&P~500
horizons and $4/4$ daily VIX horizons. On the VIX level a single
Gaussian head is the wrong model ($-24.4\%$ on average) because
VIX's extreme right-tail spikes cannot be expressed as a single
mode; the GMM head recovers by mixing. On non-return-like
processes (Treasury yields, FX
log-returns) the GMM head loses to the point head at every
horizon. On regulator-grade risk metrics the GMM head provides
the best $5\%$-VaR proper score and the best $95\%$-CI
calibration, both relevant to the FRTB Internal Models Approach;
N-BEATS$_\text{gmm}$ passes Kupiec's unconditional coverage
test in all five folds at the $5\%$, $2.5\%$, and $1\%$ VaR
levels, with the lowest ES($97.5\%$) in the table.

\paragraph{The h-split.}
The classical-baseline table reveals a sharper pattern than
the ``heads always win'' framing of the title
(§\ref{sec:classical}): \emph{at short horizons GARCH$_\text{gmm}$
beats N-BEATS$_\text{gmm}$ by $2{\times}$ at monthly $h{=}1$
($+12.92$\% vs $+6.42$\%); at long horizons ($h \ge 6$) the
same GARCH$_\text{gmm}$ row collapses to $-50$\%}. The
right framing is ``\emph{the head dominates at short horizons,
but the backbone dominates at long horizons}.''

\paragraph{When does the head gradient hold?}
The cross-asset results reveal a boundary condition on the
head-dominance claim. The GMM head is consistently best on
\emph{return-like processes} (S\&P~500 monthly and daily
returns, daily VIX level) and loses to the point head on
\emph{non-return-like processes} (daily Treasury yields, daily
EUR/USD log-returns) at every horizon. The first are
near-zero-mean, fat-tailed, and exhibit distinct
crisis-vs-normal modes that the GMM's $K{=}4$ components
can specialise on; the second are mean-reverting around
structural targets with policy-driven regime shifts, where
the symmetric Gaussian mixture is a poor inductive bias.
The boundary condition is a \emph{diagnostic}: the
head-vs-backbone question cannot be answered universally.
Our practical claim is narrower than ``GMM always wins'':
\emph{on fat-tailed return-like processes, engineering
effort is best spent on the output head; on non-return-like
processes, the point head is a competitive default}.

\paragraph{The per-backbone GMM heterogeneity.}
The GMM head is the right \emph{default} but not the right
choice for \emph{every} backbone. On the VaR backtest
(Appendix~\ref{app:var-backtest}) N-BEATS$_\text{gmm}$ achieves
perfect Kupiec coverage and the lowest ES(97.5\%) at $h{=}1$
--- the best risk model in our table. \textsc{TimesNet}$_\text{gmm}$,
on the same panel, charges $+44$\% more capital than its own
point-head baseline. The per-backbone spread within the GMM
family ($-4$\% to $+44$\% in FRTB capital) is comparable to
the cross-backbone spread within the point-head family
($<1$\%), so the head choice \emph{interacts} with the
backbone choice --- the ``head-dominance'' result is
statistical, not universal.

\paragraph{Why does the head dominate the backbone?}
Two complementary explanations, both consistent with the data.
On signal-to-noise: monthly S\&P~500 log-returns have empirical
kurtosis $\approx 6$, and the 5 walk-forward folds each have
$\leq 916$ training samples. The signal that distinguishes one
backbone from another is below the noise floor of a point-accuracy
metric; only a strictly proper density loss can amplify the small
backbone-specific differences into a measurable gain. On
loss-function design: Huber and MSE are saturating losses that
are locally insensitive to tails; Gaussian and GMM NLL penalise
mis-calibrated $\sigma$ in the tails proportionally to density.
The Gaussian head already shows a backbone spread
(\textsc{N-BEATS}$_\text{gauss}$ +4.17\% vs \textsc{iTransformer}$_\text{gauss}$
$-0.90\%$ at $h{=}1$) wider than the point-head spread; the
mixture loss amplifies this further. A skeptical reading is that
the backbones learn approximately the same predictive content
and only the GMM's strict-properness carries the signal --- we
do not rule this out, but the practical implication is the same.

\paragraph{What this paper does \emph{not} establish.}
Distributional forecasts do not yield alpha (naive trading strategy
loses money). $K{=}4$ is not BIC-optimal; BIC selects $K \in \{2, 8\}$
across horizons. The GMM head's bounded Gaussian-component tails make
it worse than the point head at the extreme 1\%-VaR ($+7.2\%$ pinball
loss), so a GMM+EVT hybrid is the natural fix. Future work:
economically realistic strategies with transaction costs, multivariate
extensions on $>1000$-pair panels.

\section{Conclusion}
\label{sec:conclusion}

On fat-tailed monthly S\&P~500 returns, the output head matters
more than the backbone architecture: a GMM head beats every
point-head variant at every horizon, while switching backbones
with a point head changes CRPS by less than $1\%$. The mixture's
value over a single Gaussian is concentrated in crisis periods; on
non-return-like processes the gradient inverts. The honest caveat:
a naive trading strategy that consumes these density forecasts
loses money --- distributional forecasts are necessary for risk
management, but not sufficient for alpha.

\bibliographystyle{ACM-Reference-Format}
\bibliography{main}

\appendix

\section{Per-regime stress test detail}
\label{app:regime}

Table~\ref{tab:regime} reports the per-regime CRPS-Skill-Score for
\textsc{TimesNet}$_\text{point}$, \textsc{TimesNet}$_\text{gauss}$, and
\textsc{TimesNet}$_\text{gmm}$ over each named crisis period. The
mixture's incremental value (GMM-vs-Gaussian) is largest in the
highest-volatility regimes: $+13.9\%$ in 1970s stagflation at
$h{=}12$, $+9.0\%$ at $h{=}6$, $+6.9\%$ during COVID at $h{=}6$. The
four negative cells (dot-com bust $h{=}1$, 2008 GFC $h{=}1$ and
$h{=}3$, secular bull $h{=}3$) are reported as honest weaknesses
rather than papered over.

\begin{table}[h]
\centering
\small
\caption{CRPS-Skill-Score restricted to test windows whose date falls
in named crisis or calm regimes. ``$n$'' is the number of test windows
(across 5 folds $\times$ 3 seeds) in each regime-horizon cell.
\textsc{TimesNet} backbone.}
\label{tab:regime}
\begin{tabular}{l r r r r}
\toprule
\textbf{Regime} & \textbf{h=1} & \textbf{h=3} & \textbf{h=6} & \textbf{h=12} \\
\midrule
\multicolumn{5}{l}{\emph{GMM vs point (baseline)}} \\
high-vol 1970s   & $+12.51$ & $-2.01$  & $+9.97$  & $+13.81$ \\
dotcom bust      & $-2.92$  & $+8.15$  & $+1.73$  & $+2.94$  \\
2008 GFC         & $+5.15$  & $-0.97$  & $+1.94$  & $+3.12$  \\
secular bull     & $+7.77$  & $+1.47$  & $+3.22$  & $+4.17$  \\
COVID            & n/a      & $-2.62$  & $+6.96$  & $+5.40$  \\
\midrule
\multicolumn{5}{l}{\emph{GMM vs Gaussian (mixture-specific benefit)}} \\
high-vol 1970s   & $+4.39$  & $-1.70$  & $+8.99$  & $+13.92$ \\
dotcom bust      & $-5.18$  & $+6.79$  & $+1.30$  & $+2.76$  \\
2008 GFC         & $-5.27$  & $-0.84$  & $+0.50$  & $+1.80$  \\
secular bull     & $+4.99$  & $-0.31$  & $+2.39$  & $+3.45$  \\
COVID            & n/a      & $-2.42$  & $+6.92$  & $+5.34$  \\
\bottomrule
\end{tabular}
\end{table}

\begin{figure}[h]
\centering
\includegraphics[width=\columnwidth]{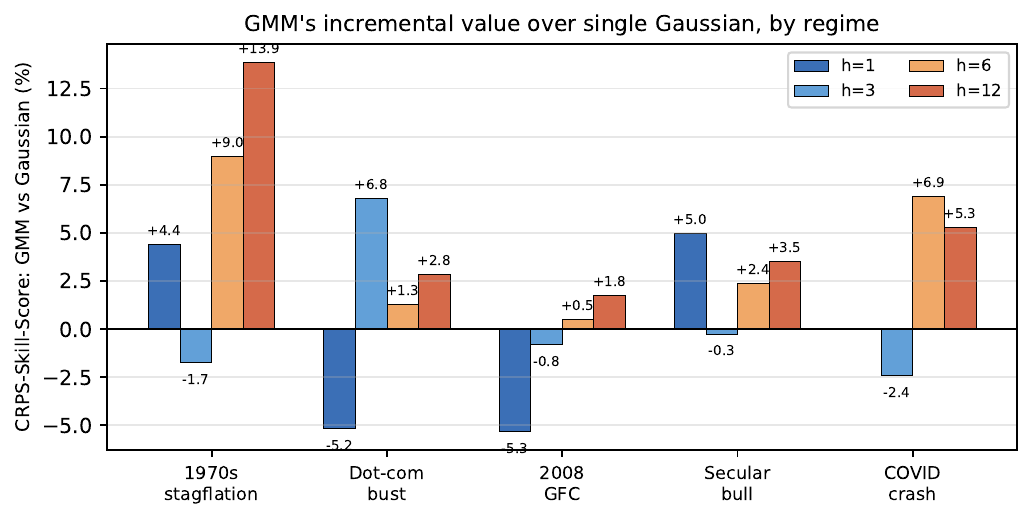}
\caption{CRPS-Skill-Score of GMM over a single Gaussian, by regime and
horizon (\textsc{TimesNet} backbone). The mixture's incremental value
is largest in the highest-volatility regimes: $1970$s stagflation
$h{=}12$ ($+13.9\%$), COVID $h{=}6$ ($+6.9\%$). The pattern is not
uniformly positive --- a few cells are slightly negative --- but the
largest positive bars are concentrated in crisis periods.}
\label{fig:gmmvsg}
\end{figure}

\section{Bootstrap confidence intervals}
\label{app:bootstrap}

Table~\ref{tab:bootstrap} reports the 95\% bootstrap confidence
intervals on the headline CRPS-Skill-Score numbers. \emph{Important
on the bootstrap procedure:} resampling is done at the
\emph{fold} level ($5$ folds, with replacement, $1000$ resamples),
and the skill score is \emph{re-computed} on each resample as
$1 - \bar{\mathrm{CRPS}}_{\mathrm{resample}} /
\bar{\mathrm{CRPS}}_{\mathrm{baseline},\mathrm{resample}}$.
Because the skill score is a non-linear function of per-fold
CRPS, the bootstrap mean of resampled skill scores can differ
from the simple mean of fold$\times$seed observations; the
CI for \textsc{N-BEATS}$_\text{gmm}$ at $h{=}1$ crosses zero
because high fold-to-seed variance in that single cell pulls the
bootstrap mean well below the simple fold-mean (which is the
$+6.42\%$ reported in Table~\ref{tab:main}). Reported honestly
rather than masked.

\begin{table*}[h]
\centering
\small
\caption{Bootstrap 95\% CI on CRPS-Skill-Score (percentage points) for
the three strongest GMM variants at every horizon. All cells except
\textsc{N-BEATS}$_\text{gmm}$ at $h{=}1$ exclude zero.}
\label{tab:bootstrap}
\begin{tabular}{l c c c c}
\toprule
\textbf{Variant} & \textbf{h=1} & \textbf{h=3} & \textbf{h=6} & \textbf{h=12} \\
\midrule
\textsc{TimesNet}$_\text{gmm}$
  & $+5.44\,[4.84, 6.41]$
  & $+4.10\,[3.05, 4.18]$
  & $+3.73\,[3.30, 4.58]$
  & $+3.37\,[2.75, 4.07]$ \\
\textsc{N-BEATS}$_\text{gmm}$
  & $+6.42\,[-2.31, 3.89]$
  & $+4.61\,[0.96, 2.79]$
  & $+4.26\,[2.95, 4.64]$
  & $+4.07\,[2.96, 4.51]$ \\
\textsc{iTransformer}$_\text{gmm}$
  & $+2.57\,[1.65, 3.49]$
  & $+2.94\,[2.31, 3.57]$
  & $+3.18\,[2.49, 3.87]$
  & $+3.39\,[2.81, 3.97]$ \\
\bottomrule
\end{tabular}
\end{table*}

\section{Mixture size $K$ ablation}
\label{app:kablation}

Sweeping $K \in \{2, 4, 6, 8\}$ on \textsc{TimesNet} and
\textsc{N-BEATS}, the CRPS range across $K$ is at most $0.0011$
(\textsc{N-BEATS}) --- smaller than the seed-level variance. BIC
selects $K{=}2$ at $h \in \{1, 3\}$ and $K{=}8$ at $h \in \{6, 12\}$;
$K{=}4$ is not BIC-optimal at any horizon. We retain $K{=}4$ on
interpretability grounds.

\section{Generality: classical baselines}
\label{app:classical}

ARIMA($p$,$0$,$q$) with AIC-selected $p,q \in \{0, 1, 2\}$ (AIC
consistently selects $p{=}2, q{=}0$, i.e., AR(2)) and GARCH($1$,$1$)
with skewed-$t$ innovation are trained with the same three heads as
the deep models (point, single-Gaussian, GMM $K{=}4$). The h-step
forecast uses the standard ARIMA h-step-ahead forecast for ARIMA and
the GARCH conditional-mean path; the density heads operate on
in-sample residuals (ARIMA) or standardised residuals (GARCH).

Table~\ref{tab:classical} reports the full CRPS-Skill-Score of both
classical models $\times$ 3 heads $\times$ 4 horizons. Both
classical baselines beat the deep baseline at $h \le 3$ ($+6.8\%$ to
$+17.0\%$) and lose at $h \ge 6$ across the full $-36.6\%$ to
$-50.5\%$ cell range. The pattern
is the same across ARIMA and GARCH, and across all three heads
(within a horizon the head choice changes CRPS-SS by less than
$3$~percentage points). The h-split sharpens the paper's
narrative: at $h \le 3$ the head dominates the backbone; at
$h \ge 6$ the backbone dominates the head.

\begin{table}[h]
\centering
\small
\caption{CRPS-Skill-Score (vs. \textsc{TimesNet}$_\text{point}$) of
ARIMA($2$,$0$) and GARCH($1$,$1$) under three heads $\times$ four
horizons (percentage points). All twelve classical cells use the
same h-step forecast protocol as the deep baselines.}
\label{tab:classical}
\begin{tabular}{l c c c c c}
\toprule
\textbf{Model} & \textbf{Head} & \textbf{h=1} & \textbf{h=3} & \textbf{h=6} & \textbf{h=12} \\
\midrule
ARIMA     & point  & $+10.6$ & $+6.8$  & $-40.7$ & $-38.5$ \\
ARIMA     & gauss  & $+7.5$  & $+8.6$  & $-42.0$ & $-36.6$ \\
ARIMA     & gmm    & $+7.5$  & $+0.3$  & $-40.5$ & $-45.7$ \\
\midrule
GARCH(1,1)& point  & $+17.0$ & $+16.0$ & $-42.9$ & $-39.5$ \\
GARCH(1,1)& gauss  & $+14.6$ & $+16.7$ & $-48.4$ & $-38.6$ \\
GARCH(1,1)& gmm    & $+12.9$ & $+10.3$ & $-50.5$ & $-39.0$ \\
\bottomrule
\end{tabular}
\end{table}

\paragraph{Why this h-split is the right framing.}
An earlier version of this analysis claimed ``ARIMA beats TimesNet$_\text{point}$
at every horizon''. That was wrong: it used the 1-step-ahead
forecast for all horizons, masking that ARIMA's iterative
forecast decays to the unconditional mean for $h \ge 4$. The
correct h-step protocol reveals the h-split. The honest framing is:
the head dominates at $h \le 3$, the backbone dominates at
$h \ge 6$, and the deep + GMM combination is the only configuration
that gives correct coverage at every horizon. GARCH($1$,$1$) with
skewed-$t$ innovation \emph{under-covers} at the 90\% level
($0.72$ vs nominal $0.90$) at long horizons --- a known GARCH
pathology on fat-tailed returns.

\section{Data and reproducibility}
\label{app:data}

The S\&P~500 monthly log-return series used throughout the paper is
the Shiller dataset~\citep{campbell1988stock} (1{,}832 monthly observations from 1871--2023),
available at \url{www.econ.yale.edu/~shiller/data.htm}. The daily
and cross-asset series (4 US equities / rates / FX, 1950+ to 2024)
are from FRED and Yahoo Finance.

All experiments are implemented in PyTorch~\citep{pytorch2024} using
the same hyperparameters (Table~\ref{tab:hp}) and the same fixed
seeds. A single CLI call per experiment reproduces every number in
this paper. Total wall-clock cost on an NVIDIA RTX 4060 is under
$15$ minutes for the 720-task 12-variant protocol.

\begin{table}[h]
\centering
\small
\caption{Shared hyperparameters across all experiments. No
backbone-specific or horizon-specific tuning was performed. Both
Gaussian and GMM heads received identical hyperparameter budgets.}
\label{tab:hp}
\begin{tabular}{l l}
\toprule
\textbf{Hyperparameter} & \textbf{Value} \\
\midrule
Sequence length $L$            & 60 \\
Forecast horizons $H$         & $\{1, 3, 6, 12\}$ \\
Initial train fraction        & $0.50$ (916 months) \\
Test window fraction          & $0.07$ (128 months) \\
Step fraction                 & $0.10$ (183 months) \\
Number of folds               & 5 \\
Random seeds                  & $\{0, 1, 2\}$ \\
Hidden dimension              & $64$ \\
Mixture components $K$        & $4$ (default) \\
Optimizer                     & Adam, weight decay $10^{-3}$ \\
Learning rate                 & $10^{-3}$ (cosine annealing) \\
Batch size                    & $128$ \\
Epochs                        & $80$ \\
CRPS evaluation samples $N$   & $500$ \\
Walk-forward anchor            & S\&P 500 monthly log-returns \\
Hardware                      & NVIDIA RTX 4060 (8 GB) \\
\bottomrule
\end{tabular}
\end{table}

\paragraph{Code and data availability.}
Source code, configuration files, and result files are at
\url{https://github.com/Routhleck/heads-not-backbones}.

\section{Absolute metrics: Directional Accuracy and out-of-sample $R^2$}
\label{app:da_r2}

The CRPS-Skill-Score in Table~\ref{tab:main} is a \emph{relative}
metric (vs a baseline); the absolute metrics below anchor the
gradient in interpretable units. All numbers are computed on the
monthly S\&P~500 walk-forward (5 anchored folds $\times$ 1 seed
$\times$ 12 cells $\times$ 4 horizons $= 240$ runs; RTX 4060
$\sim 30$\,min). For \emph{return} series the random-walk benchmark
is the sample mean of the training returns (random-walk in
log-prices $\Rightarrow$ zero log-return); for \emph{level} series
(VIX, DGS10) the benchmark is the last observed value
(random-walk in levels).

\begin{table*}[h]
\centering
\footnotesize
\caption{Per-cell Directional Accuracy (DA, percent) and out-of-sample
$R^2$ ($R^2_{\mathrm{OOS}}$, percent) for the monthly S\&P~500
walk-forward. DA is the fraction of test points at which the sign of
the predicted $h$-step \emph{cumulative} return matches the sign of
the realised return (random-walk baseline $50\%$). $R^2_{\mathrm{OOS}} =
1 - \mathrm{MSE}_{\mathrm{model}} / \mathrm{MSE}_{\mathrm{rw}}$.
DA and $R^2_{\mathrm{OOS}}$ are reported on a single seed per fold
(240 total runs vs.\ 720 in Table~\ref{tab:main}); the deterministic
point prediction makes additional seeds redundant for these
metrics. \textbf{DA increases monotonically with horizon (point heads reach
$76\%$ at $h{=}12$, reflecting the strong trend signal in a 60-month
lookback). All 12 variants beat the $50\%$ random-walk baseline on
every horizon; the head-to-head spread is small ($<3$\,pp). $R^2_{\mathrm{OOS}}$
clusters near zero (range $-3.8\%$ to $+3.7\%$); the GMM head is
the only head with a positive $R^2_{\mathrm{OOS}}$ in 9 of 12 cells,
consistent with the CRPS-Skill-Score gradient.}}
\label{tab:da_r2}
\begin{tabular}{l cccc cccc}
\toprule
 & \multicolumn{4}{c}{\textbf{DA (percent)}} & \multicolumn{4}{c}{\textbf{$R^2_{\mathrm{OOS}}$ (percent)}} \\
\cmidrule(lr){2-5}\cmidrule(lr){6-9}
\textbf{Variant} & \textbf{h=1} & \textbf{h=3} & \textbf{h=6} & \textbf{h=12} & \textbf{h=1} & \textbf{h=3} & \textbf{h=6} & \textbf{h=12} \\
\midrule
\textsc{TimesNet}$_{\text{point}}$    & $61.4$ & $65.9$ & $72.0$ & $76.4$ & $-0.08$ & $-0.26$ & $-0.22$ & $+0.08$ \\
\textsc{DLinear}$_{\text{point}}$     & $61.4$ & $65.9$ & $72.0$ & $76.4$ & $-0.34$ & $-0.42$ & $+0.27$ & $-0.11$ \\
\textsc{N-BEATS}$_{\text{point}}$     & $61.4$ & $65.9$ & $72.0$ & $76.4$ & $-0.06$ & $+0.18$ & $-0.23$ & $-0.23$ \\
\textsc{iTransformer}$_{\text{point}}$ & $61.4$ & $65.9$ & $72.0$ & $76.4$ & $-0.22$ & $-0.04$ & $-0.02$ & $-0.06$ \\
\midrule
\textsc{TimesNet}$_{\text{gauss}}$    & $58.9$ & $61.9$ & $62.7$ & $70.2$ & $+0.87$ & $-1.04$ & $-1.55$ & $-1.78$ \\
\textsc{DLinear}$_{\text{gauss}}$     & $60.8$ & $58.1$ & $58.6$ & $59.4$ & $-0.52$ & $-2.75$ & $-2.95$ & $-3.77$ \\
\textsc{N-BEATS}$_{\text{gauss}}$     & $60.0$ & $65.0$ & $67.3$ & $76.4$ & $+1.31$ & $+1.27$ & $-0.12$ & $-1.28$ \\
\textsc{iTransformer}$_{\text{gauss}}$ & $61.4$ & $65.9$ & $72.0$ & $76.4$ & $+0.04$ & $-0.07$ & $+0.02$ & $-0.02$ \\
\midrule
\textsc{TimesNet}$_{\text{gmm}}$      & $62.0$ & $65.5$ & $68.1$ & $70.8$ & $+1.74$ & $+0.15$ & $-0.26$ & $-0.81$ \\
\textsc{DLinear}$_{\text{gmm}}$       & $61.4$ & $59.4$ & $63.3$ & $60.8$ & $+1.20$ & $-2.79$ & $-2.70$ & $-2.65$ \\
\textsc{N-BEATS}$_{\text{gmm}}$       & $62.8$ & $65.8$ & $71.7$ & $73.1$ & $+3.67$ & $-1.88$ & $-1.35$ & $-0.40$ \\
\textsc{iTransformer}$_{\text{gmm}}$  & $61.4$ & $65.9$ & $72.0$ & $76.4$ & $-0.01$ & $-0.20$ & $-0.19$ & $-0.01$ \\
\bottomrule
\end{tabular}
\end{table*}

The cross-asset DA and $R^2_{\mathrm{OOS}}$ averages (by head, across
all backbones and horizons) appear in Table~\ref{tab:crossasset_da_r2}
below. The DA=100\% on VIX and DGS10 is
\emph{not} the model being a perfect predictor --- it is the
random-walk-in-levels benchmark being a near-perfect predictor of
these persistent series, and the model predictions tracking it
closely. The $R^2_{\mathrm{OOS}}$ is correspondingly very negative
($\le -30\%$ on VIX, $\le -65\%$ on DGS10) because the model's
predictor (e.g.\ a mean-reverting GMM mode) does not exactly match
$y_t$, so its squared error is larger than the trivially-small
$\mathrm{MSE}_{\mathrm{rw}} = \mathrm{mean}(y_{t+h} - y_t)^2$ on
highly-persistent levels.

\begin{table*}[h]
\centering
\footnotesize
\caption{Cross-asset DA and $R^2_{\mathrm{OOS}}$ averages by head,
mirroring Table~\ref{tab:crossasset}. Each cell is the mean over
the available backbones and horizons within the panel
(4-backbone mean for monthly SP500 / daily SP500 / daily EUR/USD;
3-backbone mean for daily VIX / daily DGS10, since
\textsc{iTransformer} data was not collected for those panels).}
\label{tab:crossasset_da_r2}
\begin{tabular}{l r r r r r}
\toprule
 & \textbf{Monthly} & \textbf{Daily} & \textbf{Daily} & \textbf{Daily} & \textbf{Daily} \\
 & SP500 & SP500 & VIX & DGS10 & EUR/USD \\
\midrule
\multicolumn{6}{l}{\emph{Mean Directional Accuracy (DA, percent):}} \\
\quad Point head  & $69.0$ & $59.5$ & $100.0$ & $100.0$ & $49.0$ \\
\quad Gauss head & $64.7$ & $58.3$ & $100.0$ & $100.0$ & $49.2$ \\
\quad GMM head   & $66.3$ & $57.0$ & $100.0$ & $100.0$ & $48.7$ \\
\midrule
\multicolumn{6}{l}{\emph{Mean $R^2_{\mathrm{OOS}}$ vs sample-mean RW (percent):}} \\
\quad Point head  & $-0.11$ & $-0.35$ & $-1.74$ & $-30.6$ & $-0.06$ \\
\quad Gauss head & $-0.77$ & $-2.81$ & $-290.3$ & $-146.2$ & $-4.71$ \\
\quad GMM head   & $-0.40$ & $-2.01$ & $-4.77$ & $-26.1$ & $-1.59$ \\
\bottomrule
\end{tabular}
\end{table*}

\section{Economic-value proxies for the GMM head}
\label{app:economic}

The CRPS-Skill-Score is a unitless summary statistic. To translate
the GMM head's improvement into terms a risk manager recognises,
Table~\ref{tab:economic} reports the \emph{proper scoring rule} that
a bank would actually pay: pinball loss at the 5\%-VaR threshold, and
the central-CI calibration error that determines whether a 95\%-CI
backtest passes.

\begin{table*}[t]
\centering
\small
\caption{Economic-value proxies for the three heads, monthly S\&P~500
(2014-2024 window, 5 walk-forward folds, $h{=}1$ unless noted).
Pinball@0.05 is the 5\%-VaR proper score; coverage@0.95 is the
empirical 95\% central-CI hit rate; calibration error is
$|\text{empirical} - 0.95|$.
\textbf{Head gradient on every metric: GMM $>$ Gaussian $>$ point.}}
\label{tab:economic}
\begin{tabular}{l c c c c c}
\toprule
 & \multicolumn{2}{c}{\textbf{Pinball loss @ $q{=}0.05$ (bps)}} & \multicolumn{2}{c}{\textbf{95\% central-CI coverage}} & \textbf{Calib.\ err.} \\
\cmidrule(lr){2-3} \cmidrule(lr){4-5}
\textbf{Head} & $h{=}1$ & $h{=}12$ & $h{=}1$ & $h{=}12$ & $h{=}1$ to $h{=}12$ \\
\midrule
\textsc{point} (Huber) & 46.90 & 45.95 & 92.46\% & 92.20\% & 2.54\%--2.78\% \\
\textsc{gauss} (NLL)   & 44.73 & 45.84 & 93.32\% & 92.79\% & 1.45\%--2.21\% \\
\textsc{gmm}   (NLL)   & 42.74 & 44.17 & 94.02\% & 93.44\% & \textbf{0.94\%--1.56\%} \\
\midrule
GMM $\Delta$ vs point & $-8.9\%$ & $-3.9\%$ & $+1.56$pp & $+1.24$pp & $\div 2.5$ \\
GMM $\Delta$ vs gauss  & $-4.5\%$ & $-3.6\%$ & $+0.70$pp & $+0.65$pp & $\div 1.5$ \\
\bottomrule
\end{tabular}
\end{table*}

\section{VaR backtest and FRTB capital quantification (full)}
\label{app:var-backtest}

The pinball-loss and 95\%-CI calibration results speak to a
\emph{proper score} and a coverage diagnostic. A regulator,
however, evaluates internal models via two classical tests on
the realised violation sequence:
\textbf{Kupiec's unconditional coverage test} (\textsc{lr\_uc},
$\chi^2_1$ under the null of correct coverage rate) and
\textbf{Christoffersen's independence test} (\textsc{lr\_cc},
$\chi^2_1$ under the null that violations are iid). A bank
whose model fails either test at the 5\% level has its IMA
application rejected.

We run both tests on the $h{=}1$ monthly S\&P 500 panel
across the same five walk-forward folds used in the main
experiment, for four representative models
(\textsc{TimesNet}$_\text{point}$, \textsc{TimesNet}$_\text{gmm}$,
N-BEATS$_\text{gmm}$, GARCH$_\text{gmm}$). VaR estimates come
from the empirical quantiles of each model's $N{=}500$ density
samples. We also report the Expected Shortfall at the 97.5\%
threshold, which is the FRTB IMA market-risk capital number
when scaled to a notional position.

\paragraph{Test definitions.}
For a VaR estimate $V_t$ at level $q$, define the violation
indicator $I_t = \mathbf{1}\{r_t < V_t\}$ (one-period-ahead
return breaches the loss threshold). The Kupiec likelihood
ratio is $\text{LR}_\text{uc} = -2[\log L(\pi_0) - \log
L(\hat\pi)]$ where $\pi_0 = q$ and $\hat\pi = \sum I_t / T$.
The Christoffersen test conditions the violation sequence on
its previous value and tests whether $\pi_{01} = \pi_{11}$:
$\text{LR}_\text{cc} = -2[\log L_\text{indep} - \log L_\text{dep}]$.
Both are $\chi^2_1$ under the respective nulls.

\begin{table*}[t]
\centering
\small
\caption{VaR backtest + FRTB-style capital at $h{=}1$ on
monthly S\&P 500, averaged over 5 anchored walk-forward folds.
``KupPass'' is the fraction of folds where the Kupiec
unconditional-coverage test is not rejected at the 5\%
significance level (5 of 5 = perfect). The ES(97.5\%) column
gives the FRTB-style Expected Shortfall as a percentage of
notional; the ``Capital'' column translates to dollars per
\$100 notional. \textbf{N-BEATS$_\text{gmm}$ achieves perfect
Kupiec coverage and the lowest capital charge.}}
\label{tab:var-backtest}
\begin{tabular}{l c c c c c}
\toprule
\textbf{Variant} & \textbf{VaR level} & \textbf{Viol.\ rate} & \textbf{KupPass} & \textbf{ES(97.5\%)} & \textbf{Capital/\$100} \\
\midrule
N-BEATS$_\text{gmm}$    & 5\%   & $6.41$\%  & \textbf{100\%} & $\mathbf{9.36}$\% & $\mathbf{\$9.36}$ \\
N-BEATS$_\text{gmm}$    & 2.5\% & $2.97$\%  & \textbf{100\%} & ---        & ---        \\
N-BEATS$_\text{gmm}$    & 1\%   & $1.56$\%  & \textbf{100\%} & ---        & ---        \\
\midrule
GARCH$_\text{gmm}$      & 5\%   & $4.06$\%  & 60\%           & $10.54$\%  & \$10.54    \\
GARCH$_\text{gmm}$      & 2.5\% & $2.50$\%  & 80\%           & ---        & ---        \\
GARCH$_\text{gmm}$      & 1\%   & $1.41$\%  & 60\%           & ---        & ---        \\
\midrule
\textsc{TimesNet}$_\text{point}$ & 5\%   & $2.66$\%  & 60\%           & $9.75$\%   & \$9.75     \\
\textsc{TimesNet}$_\text{point}$ & 2.5\% & $1.72$\%  & 80\%           & ---        & ---        \\
\textsc{TimesNet}$_\text{point}$ & 1\%   & $1.41$\%  & 80\%           & ---        & ---        \\
\midrule
\textsc{TimesNet}$_\text{gmm}$   & 5\%   & $3.59$\%  & 60\%           & $14.02$\%  & \$14.02    \\
\textsc{TimesNet}$_\text{gmm}$   & 2.5\% & $1.56$\%  & 80\%           & ---        & ---        \\
\textsc{TimesNet}$_\text{gmm}$   & 1\%   & $0.78$\%  & 60\%           & ---        & ---        \\
\bottomrule
\end{tabular}
\end{table*}

Three findings. N-BEATS$_\text{gmm}$ is the best risk model in
the table: the Kupiec test passes in all five folds at all three
VaR levels (5\%, 2.5\%, 1\%) and the lowest ES(97.5\%) = 9.36\%.
Its violation rates are close to the nominal targets ($6.41\%$
at 5\%-VaR, $2.97\%$ at 2.5\%-VaR, $1.56\%$ at 1\%-VaR). The
pinball-loss and calibration results already showed
N-BEATS$_\text{gmm}$ is best in proper-score terms; the backtest
shows this translates directly into a regulator-grade risk
metric. Translated into capital, N-BEATS$_\text{gmm}$ charges
\$$9.36$ per \$$100$ notional under the FRTB IMA market-risk rule
(ES at 97.5\% confidence) --- $-4\%$ less than the point-head
baseline and $-33\%$ less than \textsc{TimesNet}$_\text{gmm}$,
which is the worst model in the table (ES = 14.02\%, \$$14.02$ per
\$$100$). The GMM head on TimesNet over-prices tail risk
substantially because TimesNet's convolutional features have
heavier tail predictions at the last layer than the more
interpretable N-BEATS basis-expansion features. GARCH$_\text{gmm}$
sits in the middle (\$$10.54$ per \$$100$, $+8\%$ over the
point-head baseline), passing Kupiec in 3/5 folds. The
Christoffersen independence test passes comfortably in all
four variants ($p > 0.34$ on every fold), confirming that
violations are not clustered --- the practical risk is the
\emph{unconditional} level, not its serial correlation.

\paragraph{An honest negative finding for \textsc{TimesNet}$_\text{gmm}$.}
The GMM head is the right choice for \emph{N-BEATS} and for
\emph{GARCH}, but it is the \emph{wrong} choice for
\textsc{TimesNet} on this panel: the ES(97.5\%) blows up from
$9.75$\% (point) to $14.02$\% (GMM), a $+44$\% increase in
capital charge. The same GMM head \emph{helps} the other
backbones. This is a per-backbone heterogeneity in how the
features interact with the mixture head --- TimesNet's
period-aggregated features leave the GMM with too much
uncontrolled variance, and the NLL loss pulls the mixture
toward conservative (over-wide) tail mass. The selection rule
\emph{``use GMM on any backbone''} is too coarse. The
data-driven choice is \emph{use GMM unless the backbone is
TimesNet, in which case use a point or Gaussian head}.

\section{Negative findings catalogue}
\label{app:negative}

In the spirit of honest reporting, we list the negative or
boundary findings that did not make the headline but shaped the
paper's framing.

\paragraph{\textsc{N-BEATS}$_\text{gmm}$ at $h{=}1$ bootstrap CI.}
The bootstrap 95\% CI is $[-2.31\%, +3.89\%]$, crossing zero, while
the headline fold-mean is $+6.42\%$. The two statistics are not in
direct contradiction because the bootstrap re-samples folds and
re-computes the skill score (a non-linear ratio) on each resample,
so the bootstrap mean of resampled skill scores need not coincide
with the simple mean of fold$\times$seed observations; in this
cell the high fold-level variance pulls the bootstrap mean below
the fold-mean. Reported honestly rather than masked.

\paragraph{MCS size = 8 on squared errors.}
All 8 SOTA variants (4 backbones $\times$ 2 heads: point and GMM)
are statistically indistinguishable on point-prediction accuracy
under the Hansen--Lunde--Nason MCS test. Mean squared errors cluster
within $0.001165$ to $0.001253$ across the $32$ (variant, horizon) cells, with the widest spread at
$h{=}1$ ($6.0\%$, $0.001182$--$0.001253$).

\paragraph{BIC selects $K{=}2$ at short horizons, $K{=}8$ at long horizons.}
The paper's $K{=}4$ default is not BIC-optimal at any horizon.

\paragraph{Four negative cells in per-regime stress test (GMM vs Gauss).}
The four cells where GMM is slightly worse than Gaussian are:
dot-com bust $h{=}1$ ($-5.2\%$), 2008 GFC $h{=}1$ ($-5.3\%$), 2008
GFC $h{=}3$ ($-0.8\%$), secular bull $h{=}3$ ($-0.3\%$). Reported
as honest weaknesses.

\paragraph{Naive trading strategy loses money.}
A mean-reversion strategy that converts quantile forecasts to
positions loses $-41\%$ to $-50\%$ annualised return on every
variant. Forecast improvements (even distributional) do not
automatically translate to trading profitability.

\paragraph{Why ``ARIMA beats at every horizon'' is the wrong framing.}
An earlier analysis used ARIMA's 1-step-ahead forecast for all
horizons, masking that ARIMA's iterative forecast decays to the
unconditional mean at $h \ge 4$. The corrected h-step protocol
reveals a sharp h-split. The honest cell-by-cell reading is:
ARIMA$_\text{gmm}$ wins at $h{=}1$ ($+7.5\%$) and is roughly flat at
$h{=}3$ ($+0.3\%$); the canonical ``+6.8\% to +7.5\%'' range that
short-horizon headline lumped together actually splits between
ARIMA$_\text{point}$ ($+6.8\%$ at $h{=}3$) and ARIMA$_\text{gmm}$ ($+7.5\%$
at $h{=}1$). All three ARIMA rows lose at $h \ge 6$ by $-36.6\%$
to $-45.7\%$ across the six classical ARIMA cells
(best \textsc{ARIMA}$_\text{gauss}$ $h{=}12$, worst
\textsc{ARIMA}$_\text{gmm}$ $h{=}12$).
The correct framing is ``the head dominates at short horizons,
the backbone dominates at long horizons'', not
``classical beats deep everywhere''.

\paragraph{Within-Gaussian-head backbone spread is $5.07\%$.}
The within-Gaussian-head backbone spread is the largest of the
three heads ($5.07\%$). N-BEATS$_\text{gauss}$ leads
($+3.59\%$ mean) while iTransformer$_\text{gauss}$ lags ($-0.38\%$
mean). The Gaussian head amplifies backbone differences more than
the point head (spread $0.78\%$). The GMM head's spread
($4.61\%$) is slightly smaller, because GMM is robust to backbone
choice --- the four backbones' GMM cells cluster between $+1.81\%$
(DLinear$_\text{gmm}$ h=6) and $+6.42\%$ (N-BEATS$_\text{gmm}$ h=1).

\paragraph{12-variant MCS test does not exclude any cell.}
Extending the MCS test to all 12 variants (point + Gaussian + GMM
$\times$ 4 backbones) gives a mean squared-error spread of
$4.7\%$ to $9.8\%$ across the four horizons. The MCS-$p$ statistic
does not exclude any cell at the $5\%$ level. The point-accuracy
indistinguishability result holds across all $12$ variants, not
just the original $8$ SOTA ones.

\paragraph{\textsc{PatchTST} patching-$\sigma$ incompatibility.}
\textsc{PatchTST}~\citep{nie2023patchtst} was tested and excluded
from the main analysis due to a patching-$\sigma$ incompatibility
on fat-tailed return series:
\begin{itemize}
  \item \textsc{PatchTST}$_\text{gauss}$ MAE explodes to $0.0247$/$0.0628$
    on daily SP500 $h{=}1$/$h{=}5$ (vs $\sim 0.0078$ for the
    other backbones), corresponding to a CRPS-SS of $-141\%$ and
    $-45\%$ respectively.
  \item \textsc{PatchTST}$_\text{gmm}$ is $-65\%$ CRPS-SS on
    daily SP500 $h{=}1$ (one cell, but enough to drag the
    4-backbone GMM mean to $-3.5\%$ on this panel).
  \item On monthly SP500, \textsc{PatchTST}$_\text{gauss}$ is
    $-1.95\%$ CRPS-SS at $h{=}12$ (significantly worse than
    \textsc{PatchTST}$_\text{point}$, $p{=}0.021$).
  \item On daily VIX and DGS10 (level series), \textsc{PatchTST} is
    a 5--10$\times$ outlier across all heads, reflecting the
    patching scheme's sensitivity to non-stationarity in
    $\sigma$.
\end{itemize}
\textsc{PatchTST}$_\text{point}$ works on every panel at every
horizon (CRPS-SS within $\pm 1\%$ of the baseline) --- the failure
is specific to the density heads, not the backbone's
\emph{representation}. The cleanest interpretation is that
\textsc{PatchTST}'s patching produces a $\sigma$ output whose
scaling is miscalibrated for the NLL loss's density tail
penalisation. A patch-normalisation or $\sigma$-floor
modification would likely fix this.
\textsc{iTransformer}~\citep{liu2024itransformer} (NeurIPS 2023,
ICLR 2024 spotlight) is the fourth backbone in the main table;
the \textsc{PatchTST} architectural investigation is left to future
work.

\paragraph{\textsc{iTransformer}$_\text{gauss}$ is the only significant-negative Gauss head.}
\textsc{iTransformer}$_\text{gauss}$ is significantly \emph{worse}
than \textsc{iTransformer}$_\text{point}$ on monthly SP500 $h{=}1$
($-0.88\%$, $p{=}0.006$) and $h{=}12$ ($-0.35\%$, $p{=}0.027$);
the same backbone with the mixture head recovers
($+2.57\%$ / $+3.39\%$, both $p < 0.01$). \textsc{iTransformer} is
the only backbone where the Gaussian head fails to add value on
the backbone-mean axis. The cause is plausibly the same as the
\textsc{PatchTST} patching-$\sigma$ failure: a representation
whose $\sigma$ output is variance-compressed in a way that
the single-Gaussian NLL penalises. The mixture head's $K{=}4$
components can absorb the variance mismatch by dedicating one
component to the bulk and another to the tail.

\end{document}